\documentclass[11pt]{article}
\usepackage{fullpage}
\usepackage{appendix}
\usepackage[pdftex]{graphicx}
\usepackage{subfigure}
\usepackage{va}
\usepackage[margin= 1in]{geometry}

\let\oldbibliography\thebibliography
\renewcommand{\thebibliography}[1]{%
  \oldbibliography{#1}%
  \setlength{\itemsep}{0pt}%
}

\usepackage{amsfonts, amsmath, amssymb, amsthm, constants,comment}
\usepackage[usenames]{color}
\usepackage{dsfont}
\usepackage{algorithm} %ctan.org\pkg\algorithms
\usepackage{algpseudocode}
\usepackage{mathtools}

\definecolor{darkred}{RGB}{100,0,0}
\definecolor{darkgreen}{RGB}{0,100,0}
\definecolor{darkblue}{RGB}{0,0,150}

\usepackage{hyperref}
\hypersetup{colorlinks=true, linkcolor=red, citecolor=blue, urlcolor=darkgreen}
\hypersetup{colorlinks=true, linkcolor=red, citecolor=blue, urlcolor=darkgreen}
\newtheorem{remark}{Remark}

\providecommand{\scalT}[2]{\left\langle{#1},{#2}\right\rangle}

\title{\textbf{\large{ Deep Multi-Modal Learning for Audio-Visual Speech Recognition}}}

\author{\normalsize{\textsc{
Youssef Mroueh$^{\star}$ \footnote{This work was done while Youssef Mroueh was an intern in the Speech and Algorithms Group at IBM T.J Watson Research Center.}, Etienne Marcheret$^{\dagger}$, Vaibhava Goel $^{\dagger}$}}\\
\\
\small \em $\star$  Poggio Lab, CSAIL, LCSL, Massachussetts Institute of Technology \\
\small \em and Istituto Italiano di Tecnologia. \\
\small \em $\dagger$ IBM T.J Watson Research Center\\
\\
{\small \tt  ymroueh@mit.edu,etiennem@us.ibm.com,vgoel@us.ibm.com }}

\date{\today}
\begin{document}

\maketitle

\begin{abstract}
In this paper, we present methods in deep multimodal learning for fusing speech and visual modalities for Audio-Visual Automatic Speech Recognition (AV-ASR).  First, we study an approach where uni-modal deep networks are trained separately and their final hidden layers fused to obtain a joint feature space in which another deep network is built.  While the audio network alone achieves a phone error rate (PER) of $41\%$ under clean condition on the IBM large vocabulary audio-visual studio dataset, this fusion model achieves a PER of $35.83\%$ demonstrating the tremendous value of the visual channel in phone classification even in audio with high signal to noise ratio.
%This shows an advantage of the visual channel in enhancing the recognition in a clean speech scenario.
Second, we present a new deep network architecture that uses a bilinear softmax layer to account for class specific correlations between modalities.  We show that combining the posteriors from the bilinear networks with those from the fused model mentioned above results in a further significant phone error rate reduction, yielding a final PER of $34.03\%$.
\end{abstract}
%
%\{keywords}
%Audio-Visual Automatic Speech Recognition (AV-ASR), Multimodal Learning, Deep Neural Networks.
%\end{keywords}
%
\section{Introduction}
\label{sec:intro}
\indent
 Human speech perception is not only about hearing but also about seeing: our brain integrates the waveforms representing the speech information  as well as  the lips poses and motions, often called visemes, which carry important visual information about what is being said. This has been demonstrated by the so called McGurk effect \cite{mcgurk},  which shows that a voicing of \emph{ba} and a mouthing of \emph{ga} is perceived as being~\emph{da}. In the presence of noise and multiple speakers (cocktail party effect), humans rely on lip reading in order to  enhance speech recognition \cite{lipmspeaker}.
The visual information is also important in a clean speech scenario as it helps in disambiguating voices with similar acoustics \cite{clean}.\\
\indent In Audio-Visual Automatic Speech Recognition (AV-ASR), both audio recordings and  videos of the person talking are available at  training time. It is challenging to  build models that integrates both visual and audio information, and that enhance the recognition performance of the overall system. While most previous works in AV-ASR focused on enhancing the performance in the noisy case \cite{avsr,icml11_ngiam}, where the visual information can be crucial, we focus in this paper on showing that the visual information is indeed helpful even in the clean speech scenario.\\
\indent Multimodal learning consists of fusing and relating information coming from different sources, hence AV-ASR is an important multimodal problem. 
Finding correlations between different modalities, and modeling their interactions, has been addressed in various learning frameworks and has been applied to AV-ASR \cite{Gurban09informationtheoretic,Pb,Meier96adaptivebimodal,PapandreouKPM07,PapandreouKPM09,PitsikalisKPM06}. Deep Neural Networks (DNN) have shown impressive performance in both audio and visual classification tasks,  which is why we restrict ourselves to the deep multimodal learning framework \cite{Yuhas89integrationof,icml11_ngiam,Russmultimodal,socher,galen2013-deep-cca}. \\
\indent In this paper, we propose methods in deep learning to fuse modalities, and validate them on the \emph{IBM AV-ASR Large Vocabulary Studio Dataset} (Section \ref{sec:AVSR}). First we consider the  training of two networks on the audio and the visual modality separately.
Then, considering the last layer of each network  as a better feature space, and concatenating them, we train a classifier on that joint representation, and obtain gains in Phone Error Rates (PER), with respect to an audio-only trained network. We then propose a new bilinear network that accounts for correlations between modalities and allows for joint training of the two networks, we show that a committee of such bilinear networks, fused at the level of posteriors, achieves a better PER in a clean speech scenario. \\
\indent The paper is organized as follows. In Section \ref{sec:AVSR} we present the IBM AV-ASR large vocabulary studio dataset, our feature extraction pipeline for the audio and the visual channels.  Next, in Section~\ref{Bimodal}, we present results for the fusion of networks separately trained on each modality.   In Section \ref{sec:Joint} we introduce the bilinear DNN that allows for a joint training and captures correlations between the two modalities, and derive its back-propagation algorithm in Section \ref{sec:backprop}. Finally we present posterior combination of bimodal and bilinear bimodal DNNs in Section \ref{sec:Exp}.
\section{Audio-Visual Data Set \& Feature Extraction}\label{sec:AVSR}

In this Section we present the IBM AV-ASR Large Vocabulary Studio dataset, and our feature extraction pipeline.

%%, as well as results of the fusion of separately trained networks on the audio and the visual modalities.
\subsection{IBM AV-ASR Large Vocabulary Studio Dataset}
The IBM AV-ASR Large Vocabulary Studio Dataset consists of $40$ hours of audio-visual recordings from $262$ speakers.  These were carried out in clean, studio conditions.  The audio is sampled at $16$ KHz along with the video frame rate of $30$ frames per second at $704\times480$ resolution.  The vocabulary size in these recordings is $10,400$ words.  This data set was divided into a test set of $2$ hours of audio+video from $22$ speakers, with the rest used for training.

\subsection{Feature Extraction}
%% Our feature extraction follows the pipeline given in Figure \ref{fig:pipeline}. 
For the audio channel we extract 24 MFCC coefficients at 100 frames per second.  Nine consecutive frames of MFCC coefficients are stacked and projected to 40 dimensions using an LDA matrix.  Input to the audio neural network is formed by concatenating $\pm 4$ LDA frames to the central frame of interest, resulting in an audio feature vector of dimension $360$. 

For the visual channel we start by detecting the face in the image using the openCV implementation of the Viola-Jones algorithm. We then do a mouth carving by an openCV mouth detection model.  Both these utilize the ENCARA2 model as described in~\cite{encara2}.  In order to get an invariant representation to small distortions and scales we then extract level 1 and level 2 scattering coefficients \cite{Bruna} on the $64\times64$ mouth region of interest and then reduce their dimension to 60 using LDA (Linear discriminant Analysis).  In order to match the audio frame rate we replicate video frames according to audio and video time stamps.  We also add $\pm 4$ context frames to the central frame of interest, and obtain finally a visual feature vector of dimension $540$.

%% \begin{figure}[H]
%% \includegraphics[width=\linewidth]{pipeline}
%% \caption{Low Level Features Extraction for the Audio and the Visual Channels.}
%% \label{fig:pipeline}
%% \end{figure}

\subsection{Context-dependent Phoneme Targets}

Each audio+video frame is labeled with one of $1328$ targets that represent context dependent phonemes. $42$ phones in phonetic context of $\pm 2$ are clustered using decision trees down to $1328$ classes.  We measure classification error rate at the level of these $1328$ classes, this is referred to as phone error rate (PER).

\section{Uni-modal {DNNs} \& Feature Fusion}
\label{Bimodal}

In the supervised multimodal scenario, we are given a training set $S$ of $N$ labeled examples, and $C$ classes: 
$$S=\{(x^1_i,x^2_i,y_i), i=1\dots N\}, \quad y_i \in \mathcal{Y}=\{1\dots C\},$$
where $x^1_i,x^2_i$ correspond to the first and the second modality feature vectors, respectively.
We note $t_i=e_{y_i}$ the classification targets, where $\{e_{y}\}_{y\in \mathcal{Y}}$ is the canonical basis in $\mathbb{R}^C$.
Let $\rho(y|x^1,x^2)$ be the posterior probability of being in class $y$ given the two modalities $x^1$ and $x^2$. 
\noindent In a classification task, we would like to find the model that maximizes the cross-entropy  $\mathcal{E}$:
\begin{equation}
\mathcal{E}= \frac{1}{N} \sum_{i=1}^N \sum_{y=1}^C t_i^y \log \rho(y|x^1_i,x^2_i).
\label{eq:ce}
\end{equation}

\begin{figure}[H]
\begin{center}

\includegraphics[width=0.5\linewidth]{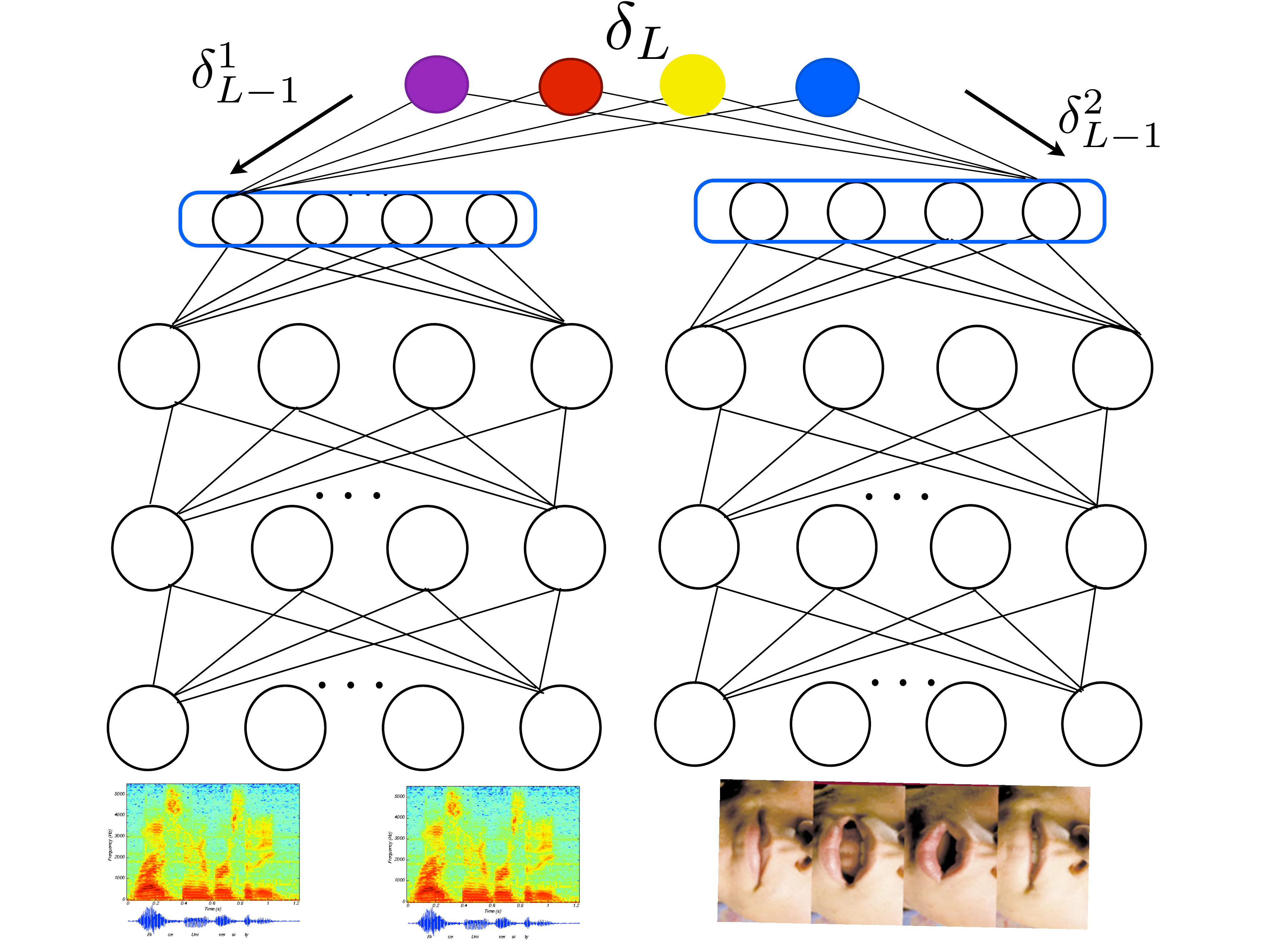}
\caption{Bimodal DNN.}
\label{fig:BiDNN}
\end{center}

\end{figure}

The first multimodal modeling approach we study is to train two separate networks $DNN_{a}$ and $DNN_{v}$ 
on the audio and the visual features, respectively.  The networks are optimized under the cross-entropy objective~\eqref{eq:ce} using the stochastic gradient descent.  We formed a joint Audio-Visual feature representation by concatenating the outputs of final hidden layers of these two networks, as shown in Figure~\ref{fig:BiDNN}.  This feature space is then kept fixed 
while a deep or a shallow (softmax only) network is trained in this fused space up to the targets.   To keep the
feature space dimension manageable, we configure the individual audio and video networks to have a low dimensional final hidden layer.
We consider for $DNN_{a}$ and $DNN_{v}$ the following architecture~$dim/1024/1024/1024/1024/1024/200/1328$, 
where $dim=360$ for $DNN_{a}$ and $dim=540$ for $DNN_{v}$.  The fused feature space dimension is $400$.  

While $DNN_{a}$ achieves a PER of $41.25\%$, $DNN_{v}$ alone achieves a PER of $69.36\%$, showing that the visual information alone carries some information but that is not enough in itself to get a low error rate.  A deep network built in the fused feature space results in a PER of $35.77\%$ while a softmax layer only in this feature space yields PER of $35.83\%$.  This substantial PER gain from joint audio-visual representation, even in clean audio conditions, demonstrates the value of visual information for the phoneme classification task.  Interestingly, the deep and the shallow fusion are roughly on par in terms of PER.  Results are summarized in the following table:
\begin{table}[H]\label{datas}
\begin{center}
  \small
    \begin{tabular}{|l|l|l| }

    \hline& PER  & Cross-Entropy \\ \hline
$DNN_{a}$ (Audio Alone) & $41.25\%$ &$1.53948848$\\
$DNN_{v}$ (Visual Alone) & $69.36\%$ & $3.24791566$\\
Bimodal (DNN Fusion ) & $ \textbf{35.77\%}$& $1.31047744$\\
Bimodal (SoftMax Fusion) & $\textbf{35.83\%}$ & $1.31077926$\\
 \hline
    \end{tabular}
    \end{center}
    \vspace*{-0.2in}
    \caption{Empirical Evaluation on the AV-ASR Studio dataset.}
    \end{table}

\section{Bilinear Deep Neural Network }\label{sec:Joint}
In the previous section the training was done separately on the two modalities, in this  section we address the joint training problem, and introduce the bilinear bimodal DNN. 

For a DNN, we note by $\sigma $ the non linearity function (sigmoid in this paper), $v_{\ell}$ the input of a unit, and $h_{\ell}$ the output of a unit in a layer $\ell$. For a layer $\ell$ we note the dimension of an  input $v_{\ell}$ by $K_{\ell}$.% for $u \in \mathbb{R},~\sigma(u)=\frac{1}{1+\exp(-u)}$.\\ 
~As shown in Figure \ref{fig:BiDNN}, we  consider two DNNs, one for each modality that we fuse at the level of the decision function. For simplicity of the exposure, we assume the same number of layers $L$ $(L_1=L_2=L)$.
For the intermediate layers, we have the standard separate networks:
$$h^j_{\ell}=\sigma( W^{j,\top}_{\ell} v^j_{\ell}+b^j_{\ell}),\quad  v^j_{\ell}=h^j_{\ell-1},\quad h^j_{0}=x^j,$$
$$W^j_{\ell} \in \mathbb{R}^{K^j_{\ell}\times K^j_{\ell+1}}, b^j_{\ell} \in \mathbb{R}^{K^j_{\ell+1}}, \ell=1\dots L-1, j \in \{1,2\}. $$
The fusion happens at the last hidden layer, where the posteriors capture the correlation between the intermediate non-linear features of the two modalities produced by the DNN layers, through a bilinear term. Let
$v^1_{L}=h^1_{L-1},~v^2_{L}=h^2_{L-1}$, the posteriors have the following form:
\begin{equation}
\rho\left(y|x^1,x^2\right)=\frac{\exp\left(v^{1,\top}_{L}W^{y}v^2_{L}+ V^{\top}_y  \left(\begin{array}{c} v^1_{L}\\v^2_{L}\end{array}\right)+b_{y}\right)}{Z},
\label{eq:bimodel}
\end{equation}
where $Z=\sum_{y'\in \mathcal{Y}} \exp\left(v^{1,\top}_{L}W^{y'}v^2_{L}+V^{\top}_{y'}  \left(\begin{array}{c} v^1_{L}\\v^2_{L}\end{array}\right)+b_{y'}\right)$, $W^{y} \in \mathbb{R}^{K^1_{L}\times K^2_{L}}, V_{y}=[V^1_y, V^2_y] \in \mathbb{R}^{(K^1_{L}+K^2_{L})}, b_y\in \mathbb{R}$ and $y\in \{1\dots C\}$.
\subsection{Factored Bilinear Softmax}
As the number of classes increases, the bilinear model becomes cumbersome computationally, and we need large training sets to get 
better estimates of the parameters. In order to decrease the computational complexity of the model, we propose the use of a factorization of the bilinear term, that is similar to the one  in \cite{G_softmax}, but is motivated in our case by Canonical Correlation Analysis (CCA) \cite{cca}:
\begin{equation}
W^{y}=U^1 \rm{diag}(w_y)U^{2,\top}, y=1\dots C,\\
\label{eq:wCCA}
\end{equation}
where $ U^1 \in \mathbb{R}^{ K^1_{L}\times F}, U^2 \in \mathbb{R}^{K^2_{L}\times F}, w_y \in \mathbb{R}^{F}$, and $\rm{diag}(w_y)$ is a diagonal matrix with $w_y$ on its diagonal . 
For numerical stability we consider $||U^j||_{F}\leq \lambda, j \in \{1,2\}$, where $\lambda$ is a regularization parameter.
We note by $F$ the dimension of the fused space, which is typically smaller than $K^1_L$ and $K^2_L$ . %This notation of the lower dimensional fused space will become clear in what follows.
 Considering the factorization in \eqref{eq:wCCA} and maximizing the cross-entropy in the bilinear model \eqref{eq:bimodel}, we have : $$\log\rho(y|x^1,x^2)=Tr( U^{1,\top}v^1_{L} v^{2,\top}_{L}U^2 \rm{diag}(w_y)) +\scalT{V^1_y}{v^1_L}+\scalT{V^2_y}{v^2_{L}}+b_y-\log(Z) .$$
For fixed weights $w_y$, learning $(U^1,U^2)$ corresponds to a class specific weighted CCA-like learning where we are looking for projections that maximize alignment between the intermediate features of the two modalities, in a discriminative way. Deep CCA of \cite{galen2013-deep-cca} shares similarities with this model.\\ 
On the other hand, for fixed  $(U^1,U^2)$, we can rewrite the log-posteriors in the following way:
$$\log(\rho(y|x^1,x^2))=\scalT{w_y}{U^{1,\top}v^1_{L}\odot U^{2,\top}v^2_{L}}+\scalT{V^1_y}{v^1_L}+\scalT{V^2_y}{v^2_{L}}+b_y-\log(Z) $$
where $\odot$ is the element-wise vector product.\\
Hence, for fixed $(U^1,U^2)$, we are learning a  linear hyperplane in the fused space of dimension $F$. 
The projection on $U^1$ and $U^2$ defines a CCA-like lower dimensional spaces, where the two modalities are maximally correlated. 
The fused space is then  defined as the element-wise vector product between two co-occurring vectors in the CCA-like lower dimensional spaces.
Hence, we can think of the last layer of the bilinear, bimodal DNN as being an ordinary softmax, having the following input $(v^1_L,v^2_L, U^{1,\top}v^1_{L}\odot U^{2,\top}v^2_{L})\in \mathbb{R}^{K^1_L+K^2_L+F}$. Therefore the decision function is learned based on the individual contributions of the modalities $v^1_L$ and $v^2_L$, as well as the joint representation produced by the fused space $U^{1,\top}v^1_{L}\odot U^{2,\top}v^2_{L}$.
\subsection{Factored Bilinear Softmax With Sharing}
When the classes we would like to predict are organized as  the leaves of a tree structure of depth two, we can further reduce the computational complexity by sharing weights between leaves having the same parent node. This is the case in AV-ASR as the $1328$ contextual phoneme states are organized as leaves of a tree, where the parent nodes correspond to $42$ different phoneme categories. In that case we share the bilinear term across leaves having the same parents.
By doing so in the case of AV-ASR, we are only taking into account the correlations between the audio and the visual channel at the phoneme level, rather than on a fine grained grid of contextual states. We can think of this sharing as a pooling operation at the phoneme level. 
More formally, assume that the label set $\mathcal{Y}$ is partitioned into $G$ non overlapping groups $\{\mathcal{Y}_g\}_{g=1\dots G}$, we assume that: 
$$W^y=W^{g}=U^1\rm{diag}(w_{g})U^{2,\top}, ~\forall~ y \in \mathcal{Y}_g , g=1\dots G.$$
Hence we reduce the number of weights to learn for the joint representation from $C \times F$ to $G\times F$.

\section{Back-propagation with the Factored Bilinear DNN with Sharing}\label{sec:backprop}
In this section we give the back-propagation algorithm and the update rules for the bilinear DNN with sharing (\emph{bi$^2$-DDN-wS}).
Recall that our classes have a tree structure with leaves $y$, and parent nodes $g$; a training example is therefore labeled by its leave label $y$ (States) as well its parent node $g$ (Phonemes), $(x^1,x^2,y,g)$, $y \in \{1\dots C\}$, and $g \in \{1\dots G\}$.
We use the notation $g(y)$ to note the group to which $y$ belongs, and we set $Root_{g(y)}=1, Root_{g}=0 , g=1\dots G ,g\neq g(y)$.\\
For the bilinear  softmax with sharing, we keep track of the errors at the level of the labels (States), as well as the groups level (Phonemes):  
\begin{align*}
\delta^k_L&= t_k -\rho(k | v^1_{L},v^2_{L}),\quad k=1\dots C, \quad \delta_{L} \in \mathbb{R}^{C\times 1}.\\
\delta^g_G&= Root_g- \sum_{k \in \mathcal{Y}_{g}}\rho(k | v^1_{L},v^2_{L}), \quad g=1\dots G, \quad \delta_{G} \in \mathbb{R}^{G\times 1}.
\end{align*}
Let $W=[w_1,\dots,w_{G}] \in \mathbb{R}^{F \times G}$, the gradients of the parameters of the bilinear softmax are given by:
\begin{align*}
\frac{\partial\mathcal{E}}{\partial W}&=\left(U^{1,\top}v^1_{L}\odot U^{2,\top}v^2_{L}\right)\delta^{\top}_{G}.\\
\frac{\partial \mathcal{E}}{\partial U^1}&=v^1_{L}v^{2,\top}_{L}U^{2}\rm{diag}(W\delta_G),
\frac{\partial \mathcal{E}}{\partial U^2}=v^2_{L}v^{1,\top}_{L}U^{1}\rm{diag}(W\delta_G).\\
\frac{\partial \mathcal{E}}{\partial V^1}&=v^{1}_{L}\delta^{\top}_{L},~
\frac{\partial \mathcal{E}}{\partial V^2}=v^{2}_{L}\delta^{\top}_{L},~
\frac{\partial \mathcal{E}}{\partial b }=\delta_{L}.
\end{align*}
For the layer right before the Bilinear softmax, we have a double projection to the first modality network (audio stream) and to the second modality network (visual stream).\\ 
We need to compute: 
\begin{align*}
W^{g}&=U^1 \rm{diag}(w_g)U^{2,\top}, g=1\dots G.\\
m^{2\rightarrow1}_{g}&=W^g v^2_{L} \quad M^{2\rightarrow 1}_{L}=[ m^{2\rightarrow 1}_{1},\dots,m^{2\rightarrow 1}_{G}] \in \mathbb{R}^{K^1_{L}\times G}.\\
m^{1\rightarrow 2}_{g}&=W^{g,\top} v^1_{L} \quad M^{1\rightarrow 2}_{L}=[ m^{1\rightarrow 2}_{1},\dots, m^{1\rightarrow 2}_{G}] \in \mathbb{R}^{K^2_{L}\times G}.
\end{align*}
Let $V^j=[V^j_1\dots V^j_C],~j\in\{1,2\}$, hence the errors we propagate to each network have the following form:
\begin{eqnarray} 
\delta^1_{L-1}&=&\frac{\partial \mathcal{E}}{\partial v^1_{L} }= M^{2\rightarrow 1}_{L}\delta_{G}+V^1\delta_L.\label{eq:errorC1}
\\ 
\delta^2_{L-1}&=&\frac{\partial \mathcal{E}}{\partial v^2_{L} }= M^{1\rightarrow 2}_{L}\delta_{G}+V^2\delta_L.\label{eq:errorC2}
\end{eqnarray}
Note that the errors now have an additional term, $M^{2\rightarrow 1}_{L}\delta_{G}$, and $M^{1\rightarrow 2}_{L}\delta_{G}$, respectively.
We can think of those terms as messages passed between networks through the bilinear term. In that way, one network influences the weights of the other one.
For the rest of the updates, it follows standard back-propagation in both networks; we give it here for completeness. Let $u^1_{\ell}=W^{1,\top}_{\ell}v^1_{\ell}+b^1_{\ell},~u^2_{\ell}=W^{2,\top}_{\ell}v^2_{\ell}+b^2_{\ell}$, then finally we have:
$\frac{\partial \mathcal{E}}{\partial W^j_{\ell}}=v^j_{\ell} (diag(\sigma'(u^j_{\ell})\delta^j_{\ell})^{\top},\frac{\partial \mathcal{E}}{\partial b^j_{\ell} }=diag(\sigma'(u^j_{\ell}))\delta^j_{\ell},$
$\delta^j_{\ell-1}=W^j_{\ell}\delta^{j}_{\ell},~j \in \{1,2\},~\ell=L-1 \dots.. 1,$
where $\delta^1_{L-1}$, and $\delta^2_{L-1}$ are given in equations \eqref{eq:errorC1} and  \eqref{eq:errorC2}. For each variable $\theta$, we have an update rule $\theta \gets \theta + \eta \frac{\partial \mathcal{E}}{\partial \theta}$, where $\eta$ is the learning rate.  For $U^1$ and $U^2$, we need to keep  control of the Frobenius norm by following the gradient step with a projection to the  Frobenius ball : $U^j \gets U^j \min(1, \frac{\lambda}{||U^j||}_{F}), j\in \{1,2\} $. 
\begin{remark} For the bilinear  softmax without sharing the update rules are similar ($\delta_G$ is replaced by $\delta_L$).
\end{remark}

\section{Combining Posteriors from Bimodal and Bilinear Bimodal Networks}
\label{sec:Exp}

We experiment with various factored \emph{bi$^2$-DNN-wS} architectures, initialized at random on the IBM AV-ASR Large Vocabulary Studio Dataset. We use the following notation for the architecture of the bilinear network: $[arch_a | arch_v | F]$, where $arch_a$ and $arch_v$ are the architectures of the audio and the visual network respectively, and $F$ is the dimension of the fused space. 
We consider architectures by increasing  complexity $Arch= [360,500,500,200,1328 | 540,500,500,200,1328 | F=200]$,
$Arch_1=[360,600,600,400,100,1328 |540,600,600,400,100,1328 |  F=100]$, and 
 $Arch_2=[360,500,500,500,500, 500,200,1328 | 540,500,500,500,500,500,200,1328 | F=200]$.\\
In all our experiments we set $\lambda=2$.  Recall that the bimodal DNN using the separate training paradigm introduced in Section \ref{Bimodal} achieves $35.83\%$ PER.  As shown in Table \ref{tab:boost}, each architecture alone does not improve on the bimodal DNN, but averaging the posteriors of the three architectures  we obtain a small gain. A gain of $1.8\%$ absolute is obtained by averaging the posteriors of the bimodal and the bilinear bimodal networks, showing that the bilinear networks have uncorrelated errors with the bimodal network. 
\begin{table}[H]
\begin{center}
  \small
    \begin{tabular}{|l|l|l| }

    \hline& PER  \\ \hline
$Arch$  & $38.89\%$ \\
$Arch_1$ & $39.01\%$ \\
$Arch_2$ & $38.36\%$\\
Bimodal & $35.83\%$\\
$Arch+Arch_1+Arch_2$ & $ \textbf{35.54\%}$\\
$Arch+Arch_1+Arch_2+\text{Bimodal}$&$\textbf{34.03\%}$\\
 \hline
    \end{tabular}
    \end{center}
    \vspace*{-0.2in}
    \caption{ Empirical evaluation on the AV-ASR Studio dataset.}
    \label{tab:boost}
    \end{table}

\section{Conclusion}
\label{sec:pagestyle}
In this paper we have studied deep multimodal learning for the task of phonetic classification from audio and visual modalities.  We demonstrate that even in clean acoustic conditions using visual channel in addition to speech results in signifiantly improved classification performance.  A bilinear bimodal DNN is introduced which leverages correlation between the audio and visual modalities, and leads to further error rate reduction.

\bibliographystyle{alpha}

\newcommand{\etalchar}[1]{$^{#1}$}

\end{document}